%% file: example_paper.tex
\documentclass{article}
\usepackage{microtype}
\usepackage{graphicx}
\usepackage{booktabs}
\usepackage{hyperref}
\usepackage[accepted]{icml2026}
\usepackage{amsmath}
\usepackage{amssymb}
\usepackage{relsize}
\usepackage{multirow}
\usepackage[capitalize,noabbrev]{cleveref}
\usepackage[section]{placeins}
\usepackage{float}

\icmltitlerunning{Characterizing Paraphrase-Induced Failures in Lean 4 Autoformalization}

\begin{document}

\twocolumn[
  \icmltitle{Characterizing Paraphrase-Induced Failures in Lean 4 Autoformalization}
  \icmlsetsymbol{equal}{*}
  \begin{icmlauthorlist}
    \icmlauthor{William Feng}{equal,yale}
    \icmlauthor{Ethan Lou}{equal,yale}
    \icmlauthor{Aryan Sharma}{equal,yale}
  \end{icmlauthorlist}
  \icmlaffiliation{yale}{Yale University, New Haven, Connecticut, USA}
  \icmlcorrespondingauthor{Ethan Lou}{ethan.lou@yale.edu}
  \icmlkeywords{autoformalization, robustness, Lean 4, theorem proving, evaluation validity}
  \vskip 0.3in
]

\printAffiliationsAndNotice{\icmlEqualContribution}

\begin{abstract}
Lean 4 autoformalization has become increasingly popular in recent years, with frontier language models and open-weight autoformalizers now producing valid formalizations of mathematical theorems. However, these evaluations often rely on single canonical phrasings of theorems and rarely probe whether outputs are robust to natural variation in inputs, while prior work has shown that semantically equivalent paraphrases often induce divergent formal outputs. We study the structure of these divergences in Lean 4 by applying deterministic paraphrase rules to datasets of undergraduate and Olympiad-level math problems. Across four frontier models and three open-weight autoformalizers, we find that paraphrase sensitivity is dominated by failures at the code-generation layer, and that these failures are typed differently by dataset. Furthermore, these patterns generalize to open-weight models, showing that state-of-the-art autoformalizers still struggle to generate valid Lean code. Our results provide a failure-mode taxonomy for autoformalization and motivate training-time interventions targeted at specific compilation failures.
\end{abstract}

\section{Introduction}
\label{sec:introduction}
Lean 4 autoformalization has matured rapidly in recent years, with frontier language models and open-weight autoformalizers~\citep{wang2025kimina, gao2025herald, xin2025deepseekproverv2} now producing valid formalizations on a meaningful fraction of standard benchmarks~\citep{azerbayev2023proofnet, zheng2022minif2f}. However, these benchmarks evaluate theorems on single natural-language phrasings, and rarely probe whether outputs are stable under input variations. Prior work has shown that semantically equivalent paraphrases of the same theorem can induce divergent formal outputs~\citep{moore2025paraphrase}, but it remains unclear how these failures are structured. Characterizing this structure would allow for targeted training-time interventions, effectively improving the reliability of generated autoformalizations.

We study this gap in Lean 4 by using a controlled suite of 60 deterministic paraphrase rules applied to ProofNet\# (a dataset of 185 formalized undergraduate-level theorems) and miniF2F (a dataset of 244 formalized Olympiad-level problems). Each rule is implemented as a regex-based surface trigger preserving semantic meaning. Rules only edit a single, named linguistic axis, e.g., conditional restructuring, concept renaming, or quantifier rephrasing --- this design lets us attribute paraphrase-driven variance to specific input features and measure consistency rates by category, in contrast to LLM-paraphrased perturbations~\citep{moore2025paraphrase}, which combine multiple linguistic axes into single edits. We evaluate four frontier models (o3, GPT-5.4, o1, o4-mini) and three open-weight 7B autoformalizers (Kimina, Herald, DeepSeek-Prover-V2) on our suite of rules, and score their paired outputs using BEq+~\citep{poiroux2025reliable} for semantic equivalence and GTED~\citep{liu2025gted} for structural similarity.

Our main finding is that paraphrase sensitivity in Lean 4 autoformalization is dominated by code-generation failures, rather than by semantic disagreement among compiling outputs. Surface consistency (i.e., pairs whose baseline and perturbed outputs are equivalent as per BEq+) rates are low (19--56\% across models), and most inconsistency arises from pairs where at least one output fails to compile. Moreover, these failures are typed differently by dataset: on ProofNet\#, models hallucinate Mathlib API identifiers that do not exist (34--50\% of failures); on miniF2F, they produce malformed syntax or fail elaboration (47--70\% of failures). These same patterns appear on the open-weight panel, indicating that state-of-the-art autoformalizers still struggle to generate valid Lean code.

Together, these results contribute (i) a failure-mode taxonomy for Lean 4 autoformalization under paraphrases, (ii) an empirical characterization of the failures across frontier and autoformalizer models, and (iii) evidence that paraphrase robustness in autoformalization is primarily a code-generation problem. Finally, they motivate training-time interventions targeted at specific compilation-failure modes, such as retrieval-augmented identifier resolution for ProofNet\#-style failures and compiler-feedback fine-tuning for miniF2F-style failures, rather than generic robustness measures.

\section{Related Work}
\label{sec:related-work}
Autoformalization refers to the translation of natural-language mathematics into a machine-verifiable formal language~\citep{wu2022autoformalization, moura2021lean4}. Recent work has substantially improved closed- and open-weight autoformalizers, including Lean-oriented systems~\citep{lu2024forml4, gao2025herald, liu2025atlas, wang2025kimina, xin2025deepseekproverv2} and retrieval-augmented approaches for premise selection over Mathlib~\citep{yang2023leandojo}. 

Despite this progress, autoformalization benchmarks~\citep{azerbayev2023proofnet, zheng2022minif2f, poiroux2025reliable} typically evaluate models on a single canonical natural-language phrasing per theorem. Recent work has shown that paraphrases in theorems significantly affect autoformalizer output~\citep{moore2025paraphrase}, yet it has not been shown whether such perturbations reflect semantic disagreement or shallower code-generation issues. Furthermore, since generated formalizations can differ syntactically from benchmark references while expressing the same theorem, evaluation also requires equivalence metrics such as BEq+ bidirectional proof search~\citep{poiroux2025reliable}, FormalAlign contrastive scoring~\citep{lu2025formalalign}, and GTED AST edit distance~\citep{liu2025gted}.

Our work relates to studies of surface robustness under semantically equivalent input variation in informal mathematical reasoning~\citep{huang2025mathperturb, hao2025putnamgap} and natural-language-to-code translation~\citep{mastropaolo2023robustness, chen2024nlperturbator}. However, autoformalization differs because correctness is compiler-verified, meaning that inconsistencies may arise not only from semantic differences but also from failures to produce syntactically valid code. Closest to our setting, \citet{moore2025paraphrase} shows that LLM-generated paraphrases of miniF2F and ProofNet can change whether outputs compile, and \citet{liu2025rethinking} documents identifier hallucination as a major failure mode in undergraduate-level autoformalization. We build upon this by using a controlled suite of deterministic paraphrases across seven models, helping us localize failures and demonstrate that sensitivity largely stems from compilation errors rather than differences in mathematical meaning.

\section{Methods}
\label{sec:methods}
\paragraph{Perturbations.}
We apply 60 named rules to theorem-statement natural language prose across 11 categories (concept rename, conditional, discourse, quantifier, verbosity, and others). Math spans (\texttt{\$...\$}) are masked so LaTeX formulas and identifiers are never altered. Every rule has an input pattern, context guards, and a textbook reference establishing meaning preservation (Appendix~\ref{sec:appendix-methodology-spec}). Of these, 14 rules apply on ProofNet\# and 37 on miniF2F. These are dataset-specific coverage counts rather than a partition of the 60-rule suite: a rule is applied only when its trigger matches and its guards pass on that corpus.\footnote{All code is available at \url{https://anonymous.4open.science/r/F232/}.} For each theorem, we refer to the original natural language phrasing as the \emph{baseline} and its rule-rewritten version as the \emph{perturbed} input. A \emph{paraphrase pair} consists of the model's baseline and perturbed Lean outputs for a given theorem and rule.

\paragraph{Datasets.}
ProofNet\# contributes 185 undergraduate theorems that often require nontrivial Mathlib API navigation to formalize, while miniF2F contributes 244 olympiad problems paired with standardized Lean formalizations that closely mirror their natural language statements. This contrast lets us test two different sources of fragility: whether models can find the correct existing definitions and lemmas in a large formal library, and whether they can preserve the intended mathematical content when the target formalization is more directly reflected in the problem statement.

\paragraph{Models.}
We evaluate four frontier models (o3, GPT-5.4, o1, o4-mini) and three open-weight 7B autoformalizers (Kimina~\citep{wang2025kimina}, Herald~\citep{gao2025herald}, and DeepSeek-Prover-V2~\citep{xin2025deepseekproverv2}). Serving details for the open-weight models are in Appendix~\ref{sec:appendix-methodology-open-weight}.

\paragraph{Evaluation metrics.}
We use BEq+~\citep{poiroux2025reliable} as the primary equivalence metric, which declares two formal statements equivalent when bidirectional proof search succeeds in both directions. As an independent structural cross-check, we use GTED~\citep{liu2025gted}, which measures tree edit distance over Lean abstract syntax trees normalized to $[0, 1]$ (where 1.0 indicates byte-identical structure). We say a paraphrase pair is \emph{compile-both} when both Lean outputs successfully elaborate, while \emph{compile rate} denotes the fraction of predictions that elaborate. Non-compiling predictions count as non-equivalent. We define \emph{surface consistency} as the fraction of paraphrase pairs whose baseline and perturbed Lean outputs are BEq+-equivalent. 

\paragraph{Implementation.}
We patch BEq+ for models that emit their own Lean preamble (Appendix~\ref{sec:appendix-methodology-beq-patch}) and report 95\% bootstrap confidence intervals throughout. For each GPT model, we issue 50 paired requests with identical inputs and observe byte-identical outputs in all cases, confirming deterministic behavior at the API layer. Open-weight models are deterministic by construction, as we serve them at temperature 0.

\begin{figure*}[!tbp]
\centering
\includegraphics[width=0.95\textwidth]{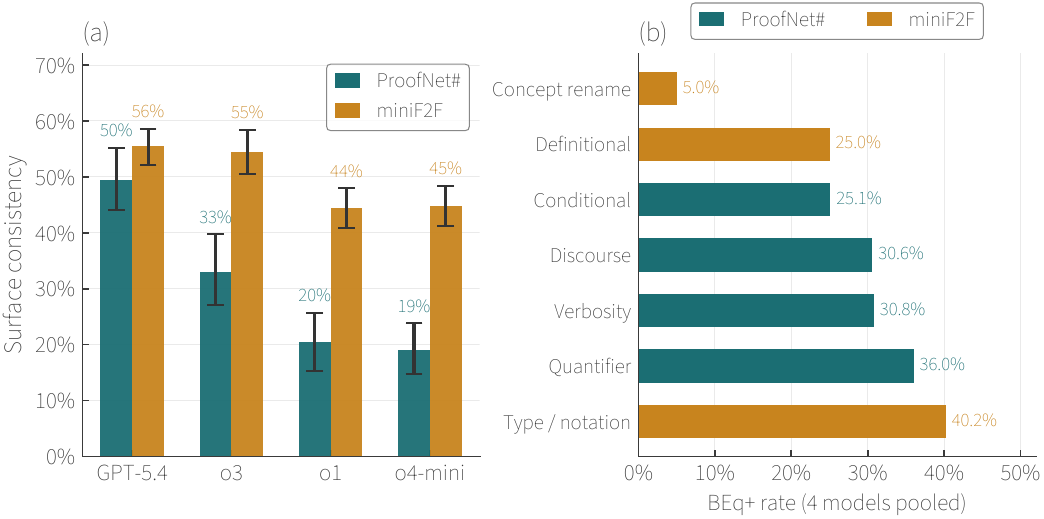}
\caption{(a) Per-model surface consistency under paraphrased inputs on
ProofNet\# (teal) and miniF2F (orange). Pair agreement ranges from
$19$--$56\%$ across the GPT panel. Error bars are $95\%$ bootstrap CIs.
(b) Per-category BEq+ surface consistency, pooled across the GPT panel,
with ProofNet\# categories (teal) and miniF2F categories (orange) shown
together.}
\label{fig:main}
\end{figure*}

\section{Results}
\label{sec:results}

\subsection{Surface Consistency}
\label{sec:results-surface}

\paragraph{Surface consistency is low across the panel.}
Across the four frontier models, surface consistency ranges from $19.1$--$49.5\%$ on ProofNet\# and $44.5$--$55.5\%$ on miniF2F (Figure~\ref{fig:main}(a)), well below the 100\% byte-identical agreement on repeated identical inputs (\S\ref{sec:methods}). Additionally, compile rates per direction are $11$--$24\%$, meaning most of the inconsistency arises from pairs where at least one Lean output fails to compile.

\paragraph{Compile-both pairs are largely equivalent.}
Among pairs where both outputs compile, BEq+-equivalence is high across the panel. Cross-checking with GTED, we find a median of at least $0.943$ across all four frontier models on ProofNet\#, with the majority of pairs at GTED $= 1.000$. This confirms that compile-both pairs are structurally near-identical, meaning the low surface consistency is not driven by semantic disagreement among compiling outputs. We provide full GTED metrics across frontier models in Appendix~\ref{sec:appendix-results-gted}.

\paragraph{BEq+ incompleteness affects a small subset.}
A small subset of BEq+ failures on our compile-both subset arises from BEq+'s known incompleteness. For example, on our \texttt{discourse\_exists\_style} perturbation, o4-mini maps ``there is an infinite number'' to a \texttt{Set.Infinite} formulation and ``there exists an infinite number'' to an explicit injective map from $\mathbb{N}$ into the quaternion solutions. These formulations are mathematically equivalent, but BEq+'s bidirectional proof search fails to bridge them. Such cases illustrate that BEq+-based non-equivalence should be read as a conservative floor, consistent with the metric's known incompleteness~\citep{liu2025rethinking, poiroux2025reliable}.

\subsection{Failure Modes Are Typed by Dataset}
\label{sec:results-failure-modes}

\paragraph{Compile failures dominate, with dataset-specific error types.}
On ProofNet\#, $34$--$50\%$ of compile failures across the models are from unknown-identifier errors: specifically, the model emits a Mathlib API name that does not exist in the pinned version, such as using \texttt{SimpleGroup} where the correct identifier is \texttt{IsSimpleGroup}. On miniF2F, $22$--$48\%$ of compile failures are syntax errors and $25$--$39\%$ are elaboration failures (Appendix~\ref{sec:appendix-results-errors}). This is because miniF2F theorems are formalized from scratch, so failures often surface as malformed Lean code rather than wrong library names. In both cases, perturbations push the model toward a made-up identifier or malformed construction, which Lean then rejects.

\paragraph{The most harmful perturbation axis is dataset-specific.}
The most destabilizing category differs across datasets (Figure~\ref{fig:main}(b)). On ProofNet\#, conditional restructuring (``if \ldots then'' $\to$ ``whenever'') causes the largest drop in surface consistency, with paired outputs equivalent in only 25.1\% of cases. On miniF2F, concept renaming is the most damaging category, with around 5\% of pairs equivalent. Per-category breakdowns for miniF2F are in Appendix~\ref{sec:appendix-results-minif2f}, while a per-rule heatmap for the frontier panel is in Appendix~\ref{sec:appendix-results-gpt-heatmap}.

We note that these differences reflect each dataset's formalization path. ProofNet\# formalizations depend on hypothesis structure, which requires navigating Mathlib's typeclass hierarchy to select the correct API entry point. Therefore, surface edits that restructure logical form change which API path the model takes. In contrast, miniF2F formalizations depend on vocabulary, because they map phrases like ``sum of the first $n$ integers'' onto specific Lean identifiers. Therefore, surface edits that rename concepts directly disrupt this lookup.

Two concrete examples of surface-induced failures are presented in Appendix~\ref{sec:appendix-results-examples}. In each case, the baseline formalization is BEq+-equivalent to the ground-truth reference, while the perturbed one is not, despite the underlying mathematical content remaining unchanged.

\subsection{Open-Weight Models Exhibit the Same Patterns}
\label{sec:results-open-weight}

\begin{figure}[!tbp]
\centering
\includegraphics[width=\columnwidth]{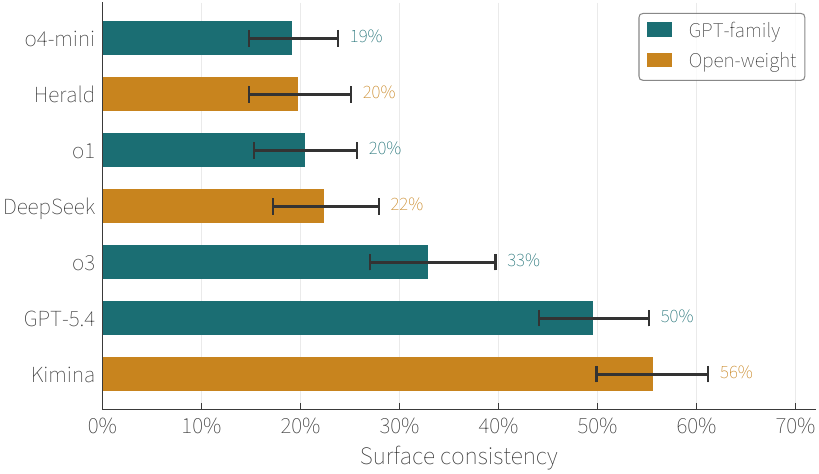}
\caption{ProofNet\# surface consistency on the 10 rules shared across both panels (95\% bootstrap confidence intervals over 5,000 iterations).}
\label{fig:open-weight}
\end{figure}

We additionally test three open-weight 7B autoformalizers (Kimina, Herald, DeepSeek-Prover-V2) on the ProofNet\# rules shared with the frontier panel and find consistency rates of $19.8$--$55.6\%$ (Figure~\ref{fig:open-weight}), overlapping the frontier-panel range of $19.1$--$49.5\%$. This indicates that the failure modes characterized above are not limited to frontier models. We note that Kimina and DeepSeek-Prover-V2 have training data overlap with ProofNet~\citep{wang2025kimina, xin2025deepseekproverv2}, so their performance partly reflects training exposure rather than generalization to unseen theorems. Nevertheless, they still show substantial paraphrase sensitivity, indicating that failure is fundamentally caused by instability in code-generation. We provide full per-rule breakdowns in Appendix~\ref{sec:appendix-results-open-weight-per-rule}.

\section{Discussion}
\label{sec:discussion}
We study paraphrase sensitivity in Lean 4 autoformalization across seven models and find that paraphrase-induced failures are dominated by the code-generation layer rather than by semantic disagreement about mathematical content. Overall, we find surface consistency to be low across the panel (19--56\% across models), with failures typed differently by dataset: on ProofNet\#, models hallucinate Mathlib identifiers (34--50\% of failures); on miniF2F, they produce malformed syntax or fail elaboration (47--70\% combined). Similar patterns appear on the open-weight panel. These findings indicate that state-of-the-art autoformalizers still struggle to consistently generate valid Lean code under realistic input variation, and that their failure modes are dataset-specific in classifiable ways. Practically, this suggests that autoformalization evaluations should report not only final equivalence or accuracy, but also where failures occur in the formalization pipeline, since identifier hallucinations, syntax errors, and elaboration failures point to different repair strategies.
\subsection{Limitations}
\label{sec:limitations}
Our equivalence claims depend on BEq+ as the primary checker. We note that BEq+ is sound but not complete by construction~\citep{liu2025rethinking, poiroux2025reliable}, and our incompleteness example shows a case where bidirectional proof search fails on a mathematically valid equivalence. For this reason, non-equivalence on BEq+ should be read as a conservative floor rather than as definitive evidence of semantic difference. Additionally, our perturbation suite primarily consists of conventional English mathematical paraphrases, while underrepresenting notational and multilingual edits. Finally, our open-weight panel is limited to three 7B autoformalizers, and whether larger or differently-trained models show the same dissociation is left for future work.

\subsection{Future Work}
\label{sec:future-work}
Our failure-mode characterization makes targeted training-time interventions feasible, since the dominant failure modes are classifiable from compiler output. Unknown-identifier errors on ProofNet\# are naturally addressable by retrieval-augmented identifier resolution, while syntax errors on miniF2F can be targeted via fine-tuning on compiler feedback. We expect interventions narrowly targeted at these specific failure modes to close the gap between paraphrase-sensitive surface consistency and paraphrase-invariant semantic capability more effectively than generic robustness measures. We view this as a natural next step.

\bibliographystyle{icml2026}
\bibliography{example_paper}

\appendix
\input{appendix}

\end{document}

%% file: appendix.tex
\raggedbottom
\setlength{\textfloatsep}{8pt plus 2pt minus 4pt}
\setlength{\floatsep}{8pt plus 2pt minus 4pt}
\setlength{\intextsep}{8pt plus 2pt minus 4pt}

\section{Additional Results}
\label{sec:appendix-results}

\subsection{Compile Rates}
\label{sec:appendix-results-compile}

Table~\ref{tab:compile-rates} reports compile rates underlying the subset analyzed in \S\ref{sec:results-surface}. The \textbf{Both} column gives the number of pairs where both baseline and perturbed predictions type-check. Baseline and perturbed compile rates are similar within each cell, meaning perturbations do not uniformly destroy compilability. Instead, the inconsistency comes from the subset of pairs where a surface change triggers a code generation failure.

\begin{table}[!htbp]
\centering
\caption{Compile rates across all frontier models and datasets. PN\# = ProofNet\#, MF2F = miniF2F.}
\label{tab:compile-rates}
\scriptsize
\begin{tabular}{llcccc}
\toprule
\textbf{Dataset} & \textbf{Model} & \textbf{N} & \textbf{Base} & \textbf{Pert} & \textbf{Both} \\
\midrule
\multirow{4}{*}{PN\#}
  & GPT-5.4 & 299 & 11.7\% & 11.4\% &  9.7\% (29) \\
  & o3      & 237 & 18.1\% & 19.0\% & 14.3\% (34) \\
  & o1      & 249 & 13.3\% & 12.0\% &  8.4\% (21) \\
  & o4-mini & 324 & 12.3\% & 12.0\% &  6.8\% (22) \\
\midrule
\multirow{4}{*}{MF2F}
  & GPT-5.4 & 827 & 15.8\% & 16.1\% & 14.6\% (121) \\
  & o3      & 582 & 23.7\% & 23.9\% & 20.6\% (120) \\
  & o1      & 723 & 24.2\% & 24.1\% & 19.9\% (144) \\
  & o4-mini & 658 & 21.7\% & 21.6\% & 16.9\% (111) \\
\bottomrule
\end{tabular}
\end{table}

\subsection{Per-Rule GPT-Panel Heatmap on ProofNet\#}
\label{sec:appendix-results-gpt-heatmap}

\begin{figure*}[!htb]
\centering
\includegraphics[width=\textwidth]{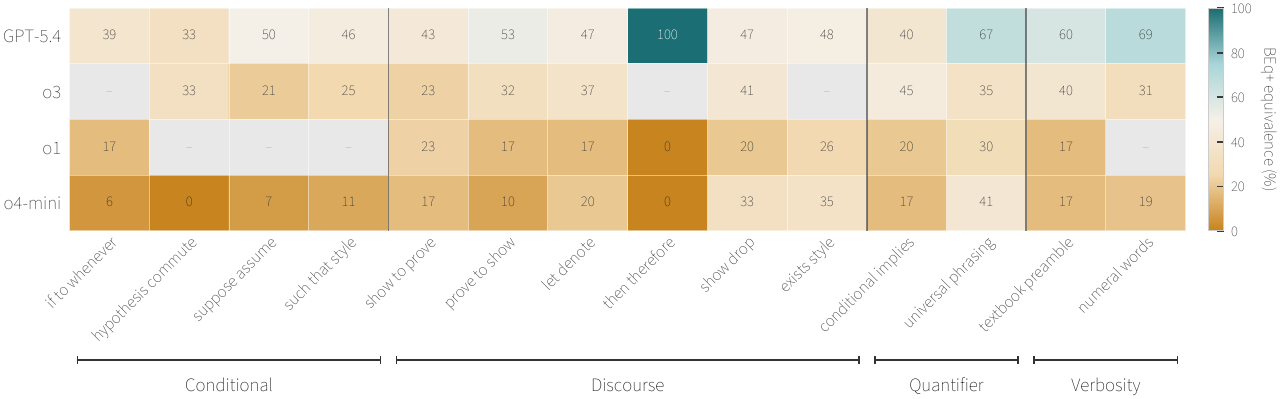}
\caption{Per-(rule, model) BEq+ equivalence on ProofNet\# (GPT panel). Rows sorted by panel-mean ascending. Dots denote cells with no paired predictions.}
\label{fig:rule-model-heatmap}
\end{figure*}

Figure~\ref{fig:rule-model-heatmap} gives the BEq+ equivalence rates for the frontier model panel on ProofNet\#. Empty cells indicate rules with zero paired predictions where both outputs compiled.

\subsection{Compile Failure Taxonomy}
\label{sec:appendix-results-errors}

We classify all non-equivalent predictions by Lean error type. The error distribution differs significantly across datasets (Table~\ref{tab:error-taxonomy}). On ProofNet\#, unknown-identifier errors account for $34$--$50\%$ of failures across models, with the model emitting a Mathlib API name that does not exist in the pinned Mathlib version (e.g., \texttt{SimpleGroup} instead of \texttt{IsSimpleGroup}). On miniF2F, syntax errors are the largest or second-largest category for every model ($22$--$48\%$), followed by elaboration failures ($25$--$39\%$). We discuss the dataset-specific structure of these failure modes in \S\ref{sec:results-failure-modes}.

\begin{table}[!htbp]
\centering
\caption{Compile error taxonomy across all non-equivalent predictions. The \textbf{Other} column aggregates type, import, and miscellaneous errors.}
\label{tab:error-taxonomy}
\scriptsize
\begin{tabular}{llrrrrr}
\toprule
\textbf{Dataset} & \textbf{Model} & \textbf{N} & \textbf{Unk.\ ID} & \textbf{Syntax} & \textbf{Elab.} & \textbf{Other} \\
\midrule
\multirow{4}{*}{PN\#}
  & GPT-5.4 & 151 & 34.4\% & 11.9\% & 23.2\% & 30.5\% \\
  & o3      & 159 & 50.3\% & 14.5\% & 14.5\% & 20.8\% \\
  & o1      & 198 & 44.9\% & 19.2\% & 10.1\% & 25.8\% \\
  & o4-mini & 262 & 46.2\% & 13.4\% & 12.6\% & 27.9\% \\
\midrule
\multirow{4}{*}{MF2F}
  & GPT-5.4 & 368 & 10.1\% & 48.4\% & 28.5\% & 13.0\% \\
  & o3      & 265 & 21.1\% & 21.9\% & 39.2\% & 17.7\% \\
  & o1      & 401 & 20.2\% & 34.2\% & 24.7\% & 20.9\% \\
  & o4-mini & 416 & 18.5\% & 31.0\% & 28.4\% & 22.1\% \\
\bottomrule
\end{tabular}
\end{table}

\subsection{Per-Category Surface Consistency on miniF2F}
\label{sec:appendix-results-minif2f}

Figure~\ref{fig:main}(b) in \S\ref{sec:results-failure-modes} reports per-category surface consistency on ProofNet\#, while Table~\ref{tab:minif2f-categories} gives the corresponding breakdown on miniF2F.

\begin{table}[!htbp]
\centering
\caption{Per-category BEq+ surface consistency on miniF2F, with 95\% bootstrap confidence intervals.}
\label{tab:minif2f-categories}
\footnotesize
\begin{tabular}{lcc}
\toprule
\textbf{Category} & \textbf{Equiv Rate} & \textbf{$N$} \\
\midrule
Concept rename    &  5.0\% [0.0, 15.0]  &     20 \\
Definitional      & 25.0\% [8.3, 41.7]  &     24 \\
Type / notation   & 40.2\% [31.6, 48.7] &    117 \\
Organization      & 44.9\% [34.7, 55.1] &     98 \\
Conditional       & 45.7\% [38.6, 52.7] &    184 \\
Verbosity         & 50.0\% [47.4, 52.7] & 1{,}368 \\
Discourse         & 52.0\% [48.8, 55.2] &    904 \\
Equivalence       & 57.1\% [28.6, 78.6] &     14 \\
Quantifier        & 58.6\% [51.0, 66.2] &    157 \\
\bottomrule
\end{tabular}
\end{table}

\subsection{Example Surface-Induced Failures}
\label{sec:appendix-results-examples}

Table~\ref{tab:examples} shows two concrete surface-induced failures. In each case, the baseline formalization is BEq+-equivalent to the ground-truth reference, while the perturbed one is not, despite the underlying mathematical content remaining unchanged.

\begin{table*}[!tbp]
\centering
\renewcommand{\arraystretch}{1.15}
\setlength{\parskip}{0.35em}
\caption{Two meaning-preserving paraphrases that flip a baseline-correct formalization to an incorrect or non-compiling Lean output. Each row shows the natural language edit and the resulting change in Lean output.}
\label{tab:examples}
\footnotesize
\begin{tabular}{@{}p{0.18\linewidth}p{0.38\linewidth}p{0.38\linewidth}@{}}
\toprule
\textbf{Rule / Model} & \textbf{NL input} & \textbf{Lean output} \\
\midrule
\addlinespace[0.5em]
\texttt{quant\_\allowbreak conditional\_\allowbreak implies}

(o1) &
\emph{Original:} Prove that if $|G|=132$ then $G$ is not simple.

\emph{Perturbed:} Prove that $|G|=132$ \textbf{implies} $G$ is not simple. &
\emph{Baseline:} \texttt{(h : Fintype.card G = 132) :\ $\neg$ IsSimpleGroup G}

\emph{Perturbed:} \texttt{(h : Fintype.card G = 132) :\ $\neg$ \textbf{SimpleGroup} G}

{\scriptsize \texttt{SimpleGroup} is not a Mathlib identifier.} \\
\midrule
\addlinespace[0.5em]
\texttt{discourse\_\allowbreak prove\_\allowbreak to\_\allowbreak show}

(o3) &
\emph{Original:} \textbf{Prove that} for all $n{>}1$, $\mathbb{Z}/n\mathbb{Z}$ is not a group under multiplication of residue classes.

\emph{Perturbed:} \textbf{We show that} for all $n{>}1$, $\mathbb{Z}/n\mathbb{Z}$ is not a group under multiplication of residue classes. &
\emph{Baseline:} \texttt{(h : 1 < n) :\ $\neg$ Nonempty (Group (ZMod n))}

\emph{Perturbed:} \texttt{(h : 1 < n) :\ $\exists$ a : ZMod n, $\forall$ b, a * b $\neq$ 1}

{\scriptsize Perturbed claim is strictly weaker: one element lacks a right-inverse vs.\ no group structure exists.} \\
\bottomrule
\end{tabular}
\end{table*}

\subsection{Per-Rule Open-Weight Surface Consistency}
\label{sec:appendix-results-open-weight-per-rule}

Table~\ref{tab:open-weight-per-rule} expands Figure~\ref{fig:open-weight} from the main paper to explore surface consistency on the ProofNet\# rules common to both the GPT and open-weight evaluations. Kimina achieves a surface consistency of 0\% on certain rules, unlike any of the other models. We leave exploring this to future work.

\begin{table}[!htbp]
\centering
\caption{Per-rule BEq+ surface consistency on the 10 ProofNet\# rules common to the GPT and open-weight evaluations.}
\label{tab:open-weight-per-rule}
\scriptsize
\setlength{\tabcolsep}{4pt}
\begin{tabular}{llcccc}
\toprule
\textbf{Cat.} & \textbf{Rule} & \textbf{N} & \textbf{Herald} & \textbf{Kimina} & \textbf{DeepSeek} \\
\midrule
\multirow{3}{*}{cond}
  & \texttt{if\_to\_whenever}  &  18 & 16.7\% &  0.0\% & 27.8\% \\
  & \texttt{such\_that\_style} &  28 & 25.0\% &  0.0\% &  0.0\% \\
  & \texttt{suppose\_assume}   &  28 & 14.3\% &  0.0\% &  0.0\% \\
\midrule
\multirow{5}{*}{disc.}
  & \texttt{exists\_style}     &  23 & 26.1\% & 91.3\% & 21.7\% \\
  & \texttt{let\_denote}       &  36 & 13.9\% & 61.1\% & 19.4\% \\
  & \texttt{prove\_to\_show}   & 121 & 20.7\% & 65.3\% & 24.8\% \\
  & \texttt{show\_drop}        &  78 & 23.1\% & 55.1\% & 26.9\% \\
  & \texttt{show\_to\_prove}   &  53 & 24.5\% & 67.9\% & 28.3\% \\
\midrule
\multirow{2}{*}{verb.}
  & \texttt{numeral\_words}    &  16 & 31.2\% & 62.5\% & 25.0\% \\
  & \texttt{textbook\_preamble}& 104 & 13.5\% & 67.3\% & 25.0\% \\
\midrule
\multicolumn{2}{l}{\textbf{Pooled}} & \textbf{505} & \textbf{19.8\%} & \textbf{55.6\%} & \textbf{22.4\%} \\
\bottomrule
\end{tabular}
\end{table}

\subsection{Panel-Wide GTED Results}
\label{sec:appendix-results-gted}

Table~\ref{tab:gted-panel} reports GTED on compile-both pairs across all frontier and open-weight models. Pooled across $401$ pairs, the median GTED is $1.000$ and the mean is $0.934$. Furthermore, eight of the eleven cells have a per-cell median of exactly $1.000$, confirming that baseline and perturbed Lean outputs are near-identical on the compile-both subset.

\begin{table}[!htbp]
\centering
\caption{Panel-wide GTED on compile-both pairs. Median TED $= 0$ in 8/11 cells, confirming structural identity is typical on compile-both pairs across the frontier and open-weight panels.}
\label{tab:gted-panel}
\scriptsize
\setlength{\tabcolsep}{4pt}
\begin{tabular}{llcccccc}
\toprule
\textbf{Model} & \textbf{Dataset} & \textbf{N} & \textbf{Median} & \textbf{p25} & \textbf{p75} & \textbf{Mean} & \textbf{count@1.0} \\
\midrule
GPT-5.4   & PN\#   & 40 & 1.000 & 1.000 & 1.000 & 0.961 & 31/40 \\
o3        & PN\#   & 40 & 1.000 & 0.885 & 1.000 & 0.901 & 22/40 \\
o1        & PN\#   & 40 & 0.943 & 0.842 & 1.000 & 0.914 & 14/40 \\
o4-mini   & PN\#   & 40 & 1.000 & 0.846 & 1.000 & 0.898 & 25/40 \\
Kimina    & PN\#   & 30 & 1.000 & 1.000 & 1.000 & 0.970 & 27/30 \\
DeepSeek  & PN\#   & 30 & 1.000 & 1.000 & 1.000 & 0.978 & 24/30 \\
Herald    & PN\#   & 21 & 1.000 & 1.000 & 1.000 & 1.000 & 21/21 \\
GPT-5.4   & MF2F   & 40 & 1.000 & 0.969 & 1.000 & 0.954 & 29/40 \\
o3        & MF2F   & 40 & 1.000 & 0.936 & 1.000 & 0.932 & 26/40 \\
o1        & MF2F   & 40 & 0.977 & 0.855 & 1.000 & 0.891 & 19/40 \\
o4-mini   & MF2F   & 40 & 1.000 & 0.958 & 1.000 & 0.927 & 28/40 \\
\midrule
\textbf{Pooled} & --- & \textbf{401} & \textbf{1.000} & \textbf{0.929} & \textbf{1.000} & \textbf{0.934} & \textbf{266/401} \\
\bottomrule
\end{tabular}
\end{table}

\section{Methodology Details}
\label{sec:method-details}

\subsection{Construction}
\label{sec:appendix-methodology-spec}

Each perturbation rule specifies an input pattern, transformation, context guards, and a reference establishing meaning preservation. Rules do not modify content inside math spans (\texttt{\$...\$}). We mask spans before matching and restore them after substitution, ensuring LaTeX formulas and variable names remain unchanged. Rules that fail their guard conditions are skipped. Table~\ref{tab:spec-examples} shows three representative rules (conditional, discourse, concept rename).

\begin{table}[!htbp]
\centering
\caption{Three representative entries from the construction-correctness spec. References: \texttt{cond\_if\_to\_whenever} is a standard logical rewrite; \texttt{discourse\_prove\_to\_show} is a proof-prose convention; \texttt{concept\_rename\_synonym} follows Dummit \& Foote, \emph{Abstract Algebra}, \S1.2.}
\label{tab:spec-examples}
\scriptsize
\setlength{\tabcolsep}{3pt}
\begin{tabular}{@{}p{0.36\linewidth}p{0.26\linewidth}p{0.31\linewidth}@{}}
\toprule
\textbf{Rule / Pattern} & \textbf{Output} & \textbf{Guard} \\
\midrule
\texttt{cond\_\allowbreak if\_\allowbreak to\_\allowbreak whenever}: ``if $P$ then $Q$'' (sentence start) &
``whenever $P$, $Q$'' &
$P$, $Q$ contain no nested conditionals \\
\midrule
\texttt{discourse\_\allowbreak prove\_\allowbreak to\_\allowbreak show}: ``Prove that'' (sentence start) &
``We show that'' (or inverse) &
Theorem-statement boundary; ignores embedded quotes \\
\midrule
\texttt{concept\_\allowbreak rename\_\allowbreak synonym}: ``abelian group'' (token, outside math) &
``commutative group'' &
Outside math spans; not \texttt{AbelianGroup} \\
\bottomrule
\end{tabular}
\end{table}

\subsection{Open-Weight Inference Setup}
\label{sec:appendix-methodology-open-weight}

We use each model's recommended prompt format. Kimina-Autoformalizer-7B uses its chat template with the system prompt ``You are an expert in mathematics and Lean 4'' and a user prompt requesting Lean 4 autoformalization. DeepSeek-Prover-V2 uses single-turn completion on the statement. Herald also uses single-turn completion and emits its own \texttt{import Mathlib} preamble (see Appendix~\ref{sec:appendix-methodology-beq-patch}). All three are run at temperature $0$ with max tokens $2048$. We extract only the \texttt{theorem ... := sorry} block from each output.

\subsection{BEq+ Wrapper Patch}
\label{sec:appendix-methodology-beq-patch}

BEq+ prepends a preamble (\texttt{import Mathlib} plus a fixed namespace) to each prediction before compilation. Herald-generated outputs already include their own preamble, and when concatenated, duplicate imports cause the Lean compiler to reject the file. For this reason, we strip the wrapper preamble before compilation; without this step, every Herald pair returns \texttt{method=failed}.